\documentclass{article}
\usepackage{spconf,amsmath,graphicx, amssymb,tabularx}
\usepackage{multirow}
\usepackage{enumitem}
\usepackage{cite}

\usepackage[small,compact]{titlesec} 
\usepackage{times}
\usepackage[small,it]{caption}

\newcolumntype{D}{>{\centering\arraybackslash}m{6ex}}
\title{Transformer-based Acoustic Modeling for Hybrid  Speech Recognition}
\name{
\begin{tabular}{c}
Yongqiang Wang$^{1}$, Abdelrahman Mohamed$^{1}$, Duc Le$^{1}$, Chunxi Liu$^{1}$, Alex Xiao$^{1}$,  \\ Jay Mahadeokar $^{1, \star}$ ,
Hongzhao Huang $^{1, \star}$, 
Andros Tjandra $^{2, \star \dagger}$, Xiaohui Zhang $^{1, \star}$, Frank Zhang $^{1, \star}$, \\
Christian Fuegen$^{1, \star}$, Geoffrey Zweig$^{1, \star}$, Michael L. Seltzer$^{1,\star}$
\end{tabular}
  \thanks{$\star$ Equal contribution; }
  \thanks{$\dagger$ Work was done when Andros was an intern at Facebook.}
}
\address{$^{1}$Facebook AI, USA \quad $^{2}$Nara Institute of Science and Technology, Japan}
\begin{document}
\ninept
\maketitle
\begin{abstract}\vspace{0.5em}
We propose and evaluate transformer-based acoustic models (AMs) for hybrid speech recognition. Several modeling choices are discussed in this work, including various positional embedding methods and an iterated loss to enable training deep transformers. We also present a preliminary study of using limited right context in transformer models, which makes it possible for streaming applications. We demonstrate that on the widely used Librispeech benchmark, our transformer-based AM outperforms the best published hybrid result by 19\% to 26\% relative when the standard $n$-gram language model (LM) is used. Combined with neural network LM for rescoring, our proposed approach achieves state-of-the-art results on Librispeech. Our findings are also confirmed on a much larger internal dataset.

\end{abstract}
\section{Introduction}
\label{sec:intro}

Since the introduction of deep learning in automatic speech recognition (ASR) \cite{hinton2012deep}, a variety of neural network architectures for acoustic modeling have been explored \cite{seide2011conversational, sak2014long, abdel2014convolutional, peddinti2015time, zhang2015feedforward}. Among them, recurrent neural networks (RNNs), especially long short-term memory (LSTM)\cite{hochreiter1997long} neural networks, are widely used, either in conventional hybrid systems (e.g., \cite{sak2014long, bahdanau2016end}), sequence-to-sequence-based (e.g. \cite{chiu2018state, park2019specaugment}) or neural-transducer-based end-to-end systems (e.g.\cite{he2019rnnt}). However, RNNs have several well-known limitations: 1) due to the vanishing or exploding gradient problem discovered in \cite{bengio1994learning}, RNNs cannot model long term temporal dependencies well; 2) the recurrence nature of RNNs makes it difficult to process speech signal in parallel. To address these issues, a variety of neural network architectures have been proposed to replace RNNs, including time delay neural networks (TDNN) \cite{peddinti2015time}, feed-forward sequential memory networks (FSMN)\cite{zhang2015feedforward}, and convolution neural networks (CNN)\cite{abdel2014convolutional, collobert2016wav2letter}, while only limited success has been achieved. 

Recently, self-attention network \cite{vaswani2017attention} has demonstrated promising results in a variety of natural language processing tasks (e.g., \cite{vaswani2017attention, devlin2018bert, radford2018improving}). Different from RNNs and CNNs, self-attention connects arbitrary pairs of positions in the input sequence directly. To forward (or backward) signals between two positions that are $n$ steps away in the input, it only needs one step to traverse the network, compared with $O(n)$ steps in RNNs and $O(\log n)$ in CNNs. Moreover, computation in self-attention can be easily parallelized.  On top of self-attention, the transformer model \cite{vaswani2017attention} leverages multi-head attention and interleaves with feed-forward layers. Self-attention and transformer models were also used for ASR, mostly in the sequence-to-sequence architecture \cite{dong2018speech, sperber2018self,zhou2018syllable} with notable exceptions of \cite{povey2018time, salazar2019self}. 

In this work, we propose and evaluate transformer-based acoustic models (AMs) for hybrid ASR. We explore several modeling choices, including methods to encode either absolute or relative positional information into the input of transformer and an iterated loss to enable training deep transformers. Though our focus in this work is to investigate the potential of transformer-based AMs without any constraint, we do explore streamable transformers and present our initial experimental results. We show that our proposed transformer-based AMs can yield significant word error rate (WER) improvement over very strong bi-directional LSTM (BLSTM) baselines, both on the widely-used Librispeech benchmark and our internal dataset. The results we obtained on Librispeech improve over the previous best hybrid WER by 19\% to 26\% when the standard 4-gram language model (LM) is used; combined with neural LM rescoring, our system achieves state-of-the-art performance on this dataset.


\section{Hybrid  Architecture}
\label{sec:hybrid}
In hybrid ASR \cite{bourlard2012connectionist}, an \emph{acoustic encoder} is used to encode an input sequence $\boldsymbol{x}_1, \cdots, \boldsymbol{x}_T$ to a sequence of high level embedding vectors $\boldsymbol{z}_1, \cdots, \boldsymbol{z}_T$. These embedding vectors are used to produce a posterior distribution of tied states of hidden Markov model (HMM), such as senone \cite{hwang1992subphonetic} or chenone \cite{le2019senones}, for each frame. These posterior distributions are then combined with other knowledge sources such as lexicons and LMs to construct a search graph. A decoder is then used to find the best hypothesis. Different neural networks can be used as the encoder: in DNN, TDNN and CNN, $\boldsymbol{z}_t$ is a function of $\boldsymbol{x}_t$ and its fixed number of neighboring frames; in uni-directional RNNs, $\boldsymbol{z}_t$ is a function of $\boldsymbol{x}_1$ to $\boldsymbol{x}_t$, while in bi-directional RNNs, $\boldsymbol{z}_t$ is a function of the entire input sequence. 

Though compared with the sequence-to-sequence or neural transducer architecture, the hybrid approach is admittedly less appealing as it is 
not end-to-end trained, it is still the best performing system for authors' practical problems. It also has the advantage that it can be easily integrated with other knowledge sources (e.g., personalized lexicon) that may not be available during training. In this work, we aim to leverage the transformer to improve hybrid acoustic modeling.

\section{Acoustic Modeling Using Transformer}
\label{sec:am}
In this section, we first briefly review the transformer network and discuss various modeling choices when using the transformer as the acoustic encoder. Relation to other works is also discussed in Section \ref{sec:relation}.

\subsection{Self-Attention and Multi-Head Attention}
Self-attention first computes the attention distribution over the input sequence using dot-product attention, i.e., for every $\boldsymbol{x}_t \in \mathbb{R}^{d_i}$, a distribution $\boldsymbol{\alpha}_{t}$ is obtained by: 
\begin{align}
    \alpha_{t\tau} = \frac{
    \exp(\beta \cdot\boldsymbol{x}_t^{\sf T}\mathbf{W}_{\rm q}^{\sf T} \mathbf{W}_{\rm k}\boldsymbol{x}_\tau )
    }{
    \sum_{\tau'} \exp(\beta \cdot\boldsymbol{x}_t^{\sf T}\mathbf{W}_{\rm q}^{\sf T} \mathbf{W}_{\rm k}\boldsymbol{x}_{\tau'} )}
\end{align}
where $\mathbf{W}_{\rm q}, \mathbf{W}_{\rm k} \in \mathbb{R}^{d_k \times d_i}$ transforms $\boldsymbol{x}_t$ to 
\emph{query} and \emph{key} space, $\beta = \frac{1}{\sqrt{d}_i}$ is a scaling factor. Note that for language modeling, the dot-products between the current position and future positions are masked to $-\infty$ to prevent future information leaking to the current embedding. Though for acoustic modeling, it is possible to attend to the entire sequence, in many applications, we only attend to limited right context frames to enable streaming processing of speech signals (i.e., dot-product between $t$ and $\tau, \tau > t + R$ is masked to $-\infty$). Given $\boldsymbol{\alpha}_t$, the output embedding of self-attention is obtained via:
\begin{align}
    \boldsymbol{z}_t = \sum_{\tau} \mathrm{Dropout}(\alpha_{t\tau}) \cdot \mathbf{W}_{\rm v} \boldsymbol{x}_\tau
\end{align}
where $\mathbf{W}_{\rm v} \in \mathbb{R}^{d_v \times d_i}$ maps the input vectors to \emph{value} space.

Self-attention is often combined with multi-head attention (MHA), where $h$ self-attention heads are applied individually on the input sequences, and the output of each head are concatenated and linearly transformed to a common space, i.e., 
\begin{align}
    \boldsymbol{z}_t = \mathbf{W}_{\rm o}\left( 
    \begin{tabular}{c}
        ... \\
        $\sum_{\tau} \mathrm{Dropout}(\alpha_{t\tau}^{(i)}) \cdot \mathbf{W}_{\rm v}^{(i)} \boldsymbol{x}_\tau$  \\
         ... 
    \end{tabular}
    \right)
\end{align}
where $\mathbf{W}_{\rm o} \in \mathbb{R}^{d_i \times hd_{\rm v}}$, $\alpha_{t\tau}^{(i)}$ and $\mathbf{W}^{(i)}_{\rm v}$ are the attention weights and the value matrix of the $i$-th head. 

\subsection{Architecture of Transformer}
\begin{figure}[hhh]
    \centering
    \includegraphics[scale=0.09]{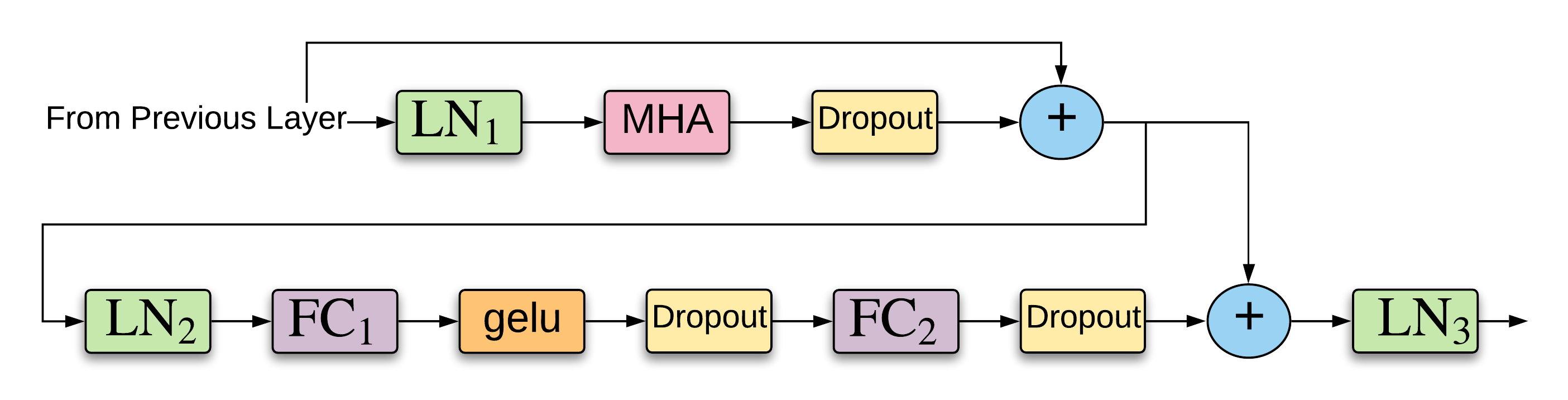}
    \caption{Architecture of one transformer layer. ``LN" means layer normalization \cite{lei2016layer}; ``FC" means fully connected linear transformation; ``gelu" means the gelu nonlinear activation \cite{hendrycks2016gaussian}.\vspace{1em}}
    \label{fig:transformer_layer}
\end{figure}

In addition to the MHA sub-layer, each transformer layer contains a fully-connected feed-forward network (FFN), which is composed by two linear transformations and a nonlinear activation function in between.
The FFN network is applied to each position in the sequence separately and identically. To allow stacking many transformer layer together, residual connections are added to the MHA and FFN sub-layers. Dropouts are also applied after MHA and linear transformation as a form of regularization. Figure \ref{fig:transformer_layer} summarizes the architecture of one transformer layer. Note 
that different from \cite{vaswani2017attention}, layer normalization \cite{lei2016layer} is applied before MHA and FFN and the third layer normalization ($\mathrm{LN}_3$ in Figure \ref{fig:transformer_layer}) is necessary to prevent bypassing the transformer layer entirely. Note, following \cite{devlin2018bert}, we use ``gelu" non-linearity \cite{hendrycks2016gaussian} in the FFN network.

\subsection{Positional Embedding}
One obvious limitation of the transformer layer is that the output is invariant to the input order permutation, i.e., for any permutation $\pi$ applied on the input sequence $\boldsymbol{x}_1, \cdots, \boldsymbol{x}_T$, the output of the transformer layer can be obtained by applying the same permutation $\pi$ on $\boldsymbol{z}_1, \cdots, \boldsymbol{z}_T$.  This means that transformer does not model the order of the input sequence. In the original transformer work \cite{vaswani2017attention}, this is solved by injecting information about absolute positions into the input sequence via sinusoid positional embeddings. We argued that different from NLP applications, relative position could be more useful for speech signals. In this work, we compare a few ways to encode positional information into the input of transformer: 

\begin{itemize}
    
    \item \emph{Sinusoid positional embedding:} a sinusoid positional embedding $\boldsymbol{p}_t$ is added to $\boldsymbol{x}_t$, where 
    the $i$-th element of $\boldsymbol{p}_t$ is $\sin((t/10000)^{i/d_i})$ for even $i$ and $\cos((t/10000)^{(i-1)/d_i}$ for odd $i$. This encodes \emph{absolute positional} information;
    \item \emph{Frame stacking:} a simple way to break the permutation invariance is to stack $n$ contextual vectors together, i.e., $\overline{\boldsymbol{x}}_t=(\boldsymbol{x}_t^{\sf T}, \boldsymbol{x}_{t+1}^{\sf T}, \cdots, \boldsymbol{x}_{t+n-1}^{\sf T})^{\sf T}$. This encodes the \emph{relative positional} information; 
    \item \emph{Convolutional  embedding:} inspired by \cite{mohamed2019transformers}, we use 2D convolutional layers to implicitly encode the relative positional information. Convolutional embedding implicitly performs \emph{frame stacking} as well as learns useful short-range spectral-temporal patterns \cite{zhang2017very}.
\end{itemize}

\subsection{Training Deep Transformers}
Transformer layers can be stacked many times to form a very deep network. In our initial experiments, we found better accuracies can be obtained by deeper networks. However, after stacking many layers, it becomes difficult to train and often gets stuck in a bad local optimum. To enable training deep transformer, we used iterated loss 
\cite{Andros2019}, in which output of some intermediate transformer layers is also used to calculate auxiliary cross entropy losses. These auxiliary losses are interpolated to make the final loss function. Note that intermediate-layer-specific parameters (e.g., the linear transformation before the softmax operation) are discarded after training. 

\subsection{Relation to Other Works}
\label{sec:relation}
The original transformer paper \cite{vaswani2017attention} proposed to use self-attention and cross-attention to replace the recurrence in encoder and decoder in a sequence-to-sequence model. Since we focus on hybrid speech recognition, we only use self-attention to replace the RNNs in the acoustic encoder in this work.

Self-attention based acoustic modeling has been explored in the past. In \cite{povey2018time}, self-attention is modified to attend to a fixed number of left and right context frames, and only one attention layer was used. By comparison, in our work attention heads attend to all the past frames, and we use both self-attention and FFN networks with a very deep structure, which is critical to achieve a good model accuracy. In \cite{karita2019comparative}, transformers are compared with RNNs in the sequence-to-sequence architecture. In \cite{sperber2018self}, various positional embedding methods were investigated for a sequence-to-sequence model, where it is found that replacing the FFN network with a LSTM layer to make the self-attention layer position aware yielded better performance. Following \cite{mohamed2019transformers}, we use convolution layers as pre-processors for the transformer layer's input and compare it with other positional encoding methods in Section \ref{ssec:positional_embedding}. In \cite{al2019character}, a loss function similar to the iterated loss is used to enable training very deep transformers for character-level LMs; we demonstrate that it is also crucial for training deep transformer-based AMs.

\section{Experiments}
\label{sec:exp}
To evaluate the effectiveness of the proposed transformer-based acoustic model, we first perform experiments on the Librispeech corpus \cite{panayotov2015librispeech}. This corpus contains about 960 hours of read speech data for training, and 4 development and test sets (\texttt{\{dev, test\} - \{clean,other\}}), where \texttt{other} sets are more acoustic challenging. No segmentation is performed for these test sets. The standard 4-gram language model (LM) with a 200K vocabulary is used for all first-pass decoding.


\subsection{Experiment Setups}

Following \cite{le2019senones}, we use context- and position-dependent graphemes (i.e., \textit{chenones}) in all experiments. We bootstrap our HMM-GMM system using the standard Kaldi \cite{Povey_ASRU2011} Librispeech recipe. We use 1-state HMM topology with fixed self-loop and forward transition probability (both 0.5). 80-dimensional log Mel-filter bank features are extracted with a 10ms frame shift. A reduced 20ms frame rate is achieved either by stacking-and-striding 2 consecutive frames or by a stride-2 pooling in the convolution layer if it is used. We found that this not only reduces the computation but also slightly improves the recognition accuracy. Speed perturbation \cite{ko2015audio} and \emph{SpecAugment} \cite{park2019specaugment} (LD policy without time warping) are used. We focus on cross-entropy (CE) trained models and only selectively perform sMBR \cite{vesely2013sequence} training on top of the best CE setup. 

Neural network training is performed using an in-house developed speech extension of the PyTorch-based \emph{fairseq}\cite{ott2019fairseq} toolkit. 
Adam optimizer \cite{kingma2014adam} is used in all experiments; the learning rate linearly warms up from 1e-5 to 1e-3 in the first 8000 iterations and stays at 1e-3 during the rest of training. We mainly compare full-context transformer with BLSTM in this work though we do have an initial investigation of transformers using limited right context. Dropout is used in all experiments: 0.1 for transformer and 0.2 for BLSTM. To improve training throughput, our batch size is dynamically determined so that we can occupy as much GPU memory as possible. For most of the experiments in this work, a batch contains around 10,000 to 20,000 frames, including padding frames. We train models using 32 Nvidia P100 GPUs for at most 100 epochs; training is usually done within 4 days. We did not perform thorough architecture searches for either transformer or BLSTM. For transformers, we mainly use a 12-layer transformer architecture with $d_i = 768$: per-head dimension is always 64 and the FFN dimension is always set to $4 d_i$. This model has about 90M parameters. For BLSTMs, we follow \cite{le2019senones} and consider two architectures, a 5-layer BLSTM with 800 units per layer per direction (about 94M parameters), and a 6-layer BLSTM with 1000 units (about 163M parameters)\footnote{We did not obtain further WER improvements by increasing number of parameters in BLSTM beyond 163M.}. 

Training transformers requires some tricks. Due to the quadratically growing computation cost with respect to the input sequence length, we segment the training utterances into segments that are not longer than 10 seconds \footnote{
This is achieved by aligning audio against the reference using an existing latency-controlled BLSTM acoustic model. }. Though this creates a mismatch between training and testing, preliminary results show that training on shorter segments not only increases the training throughput but also helps the final WERs. We also found that transformers are more prone to over-fitting, thus require some regularization. We found \emph{SpecAugment}\cite{park2019specaugment} is effective: without it, WER starts to increase after only 3 epochs, while WER continues to improve during training with \emph{SpecAugment}.

A fully-optimized, static 4-gram decoding graph is built using Kaldi. This decoding graph is used for first-pass decoding and n-best generation for neural LM rescoring. Test set WERs are obtained using the best model based on WER on the development set\footnote{We also average the last 10 epoch checkpoints to form an extra candidate.}. 
Following \cite{luscher2019rwth}, the best checkpoints for \texttt{test-clean} and \texttt{test-other} are selected separately on the corresponding development sets \footnote{This is only to follow the same experimental protocol set by the prior work in \cite{luscher2019rwth} -- most of the experimental results on both test sets, including the best WERs we reported in Table 4, are actually achieved by the same model.}

\subsection{Effect of Positional Embedding}
\label{ssec:positional_embedding}
In the first set experiment, we investigate the effect of four positional embeddings (PE) methods for transformer-based acoustic models. In the first method, we stack-and-stride every 2 frames: it does not break the permutation invariance in transformers, thus denoted as \emph{None}. In the second method, the \emph{Sinusoid} PE proposed in the original transformer paper \cite{vaswani2017attention}, which encodes the \emph{absolute} positional information, is used. In the third method, \emph{Frame Stacking}, we stack the current frame and next 8 future frames followed by a stride-2 sampling to form a new input sequence to transformers. Note that since the stacked frames are partially overlapped with its neighboring stacked frames, the permutation invariance no longer holds. This method encodes \emph{relative} positional information. In the fourth method, \emph{Convolution}, we use two VGG blocks \cite{simonyan2014very} beneath transformer layers: each VGG block contains 2 consecutive convolution layers with a 3-by-3 kernel followed by a ReLu non-linearity and a pooling layer; 32 channels are used in the convolution layer of the first VGG block and increase to 64 for the second block. Max-pooling is performed at a 2-by-2 grid, with stride 2 in the first block and 1 in the second block. For an input sequence of 80-dim feature vector at a 10ms rate, this VGG network produces a 2560-dim feature vector sequence at a 20ms rate. Note that the perception field of each feature vector output by the VGG network consists of 80ms left-context and 80ms right context, the same right context length as \emph{Frame Stacking}. A linear projection is used to project the feature vector to the dimension accepted by transformers, in this case, 768.  
\vspace{-1em}

\begin{table}[hhtb]
    \centering
    \caption{Effect of Positional Embeddings (PE) for Transformer.}
    \begin{tabular}{|c||cc|}
    \hline
    PE     &  \rm{test-clean} & \rm{test-other}\\
    \hline\hline
    \emph{None}   & 3.11  & 6.94 \\
    \emph{Sinusoid} & 3.13 & 6.67 \\
    \emph{Frame Stacking} & 3.04  & 6.64 \\
    \emph{Convolution} & 2.87 & 6.46 \\
    \hline
    \end{tabular}
    \label{tab:pe}
\end{table}
\vspace{-0.5em}


\subsection{Transformer vs. BLSTM} 

In the second set of experiments, we compare the transformer architecture with BLSTM. 
For a fair comparison, we try to build transformer and BLSTM-based models using similar number of parameters. First we compare a BLSTM model, \textit{BLSTM(800, 5)}, i.e., 5 layers with 800 hidden units per layer per direction, with the transformer model in row 3, Table \ref{tab:pe}, dubbed \textit{Trf-FS} since it uses \emph{Frame Stacking}. To be able to compare our best performing transformer-based model with \emph{Convolution} PE, we combine the same VGG blocks in row 4, Table \ref{tab:pe} with BLSTM, producing \textit{vggBLSTM(800, 5)}. Lastly, with about 163M parameters, we build the largest vggBLSTM model, \textit{vggBLSTM(1000,6)}. To match the number of parameters of this model, we increase the number of transformer layers from 12 to 20. As shown in Table \ref{tab:arch}, transformer-based models consistently outperform BLSTM-based models by 2--4\% on \texttt{test-clean} and 7--11\% on \texttt{test-other}. 

\begin{table}[htb]
    \centering
    \caption{Architecture comparison on the Librispeech benchmark}
    \begin{tabular}{|c|c|cc|}
    \hline
    Model Arch     & \#Params (M) & \rm{test-clean} & \rm{test-other} \\
    \hline\hline
    BLSTM (800,5)  & 79 & 3.11 & 7.44 \\
    Trf-FS (768,12)   & 91 & 3.04 & 6.64 \\
    \hline
    vggBLSTM (800,5) & 95 & 2.99 & 6.95 \\
    vggTrf. (768,12) & 93 & 2.87 & 6.46 \\
    \hline
    vggBLSTM (1000,6) & 163 & 2.86 & 6.63 \\
    vggTrf. (768, 20) & 149 & 2.77 & 6.10 \\
    \hline
    \end{tabular}
    \label{tab:arch}
\end{table}

\subsection{Effect of Iterated Loss}\vspace{-0.5em}

Table \ref{tab:arch} shows that simply increasing the depth of transformers to 20 layers, we obtained about 5.5\% relative WER reduction (6.10 vs. 6.46). Inspired by this, we try to increase the number of transformer layers further. To make the model size manageable, we use a smaller embedding dimension, 512, for deep transformer models. Our initial attempt was not successful; deep transformer models (deeper than 20 layers) often got stuck in training and made little progress for a long time. We solved the problem with the iterated loss used in \cite{Andros2019}: the output embeddings of the 6/12/18-th transformer layers are non-linearly transformed (projected to a 256-dimensional space with a linear transformation followed by a Relu non-linearity) and auxiliary CE losses are calculated separately. These additional CE losses are interpolated with the original CE loss with a 0.3 weight. With this iterated loss, we were able to train a 24-layer transformer model with only 81M model parameters in decoding\footnote{There are 6M extra parameters only used in training.} and obtain a 7\% and 13\% WER reduction on \texttt{test-clean} and \texttt{test-other}, respectively, over the \textit{vggTrf(768, 12)} baseline.

\begin{table}[htb]
    \centering
    \caption{Using iterated loss to train deep transformer models.}
    \begin{tabular}{|c||c|cc|}
    \hline
    Model Arch  & Iter Loss & \rm{test-clean} & \rm{test-other} \\
    \hline\hline
    vggTrf. (768, 12)   &  N & 2.87  & 6.46  \\
    (Params: 93M)    &  Y & 2.77 &  6.10 \\
    \hline
    vggTrf. (512, 24)   & N & \multicolumn{2}{c|}{not converged}  \\
    (Params: 81M) & Y & 2.66 & 5.64 \\
    \hline
    \end{tabular}
    \label{tab:iter_loss}
\end{table}

On top of this \textit{vggTrf(512, 24)} model, we further perform sMBR training and it slightly improves to 2.60\% and 5.59\% on \texttt{test-clean} and \texttt{test-other}. We compare our results with some published state-of-the-art systems on Librispeech in Table \ref{tab:comp}: when the standard 4-gram LM is used in decoding, our system achieves 19\% and 26\% WER  reduction on \texttt{test-clean} and \texttt{test-other} respectively, over previous best 4-gram only hybrid system \cite{le2019senones}\footnote{Note that \cite{le2019senones} used LC-BLSTM \cite{zhang2016highway} instead of full-context BLSTM.}. We also built a transformer LM similar to the setup in \cite{radford2018improving} on the 800M text tokens provided by the Librispeech benchmark and performed n-best rescoring on the first pass decoding output. To the best of our knowledge, our final WERs (2.26/4.85) are state-of-the-art results on this widely used benchmark.
\vspace{-0.5em}

\begin{table}[htb]
    \centering
    \caption{Comparison with previous best results on Librispeech. ``4g" means the stand 4-gram LM is used; ``NNLM" means a neural LM is used.
    }
    \begin{tabular}{|c|c|c|DD|}
    \hline
    Arch. & System & LM & test-clean & test-other \\
    \hline\hline
     \multirow{2}{*}{LAS} & Park et al.\cite{park2019specaugment} & NNLM + 4g & 2.5 & 5.8 \\
        & Karita et al. \cite{karita2019comparative} & NNLM & 2.6 & 5.7 \\
    \hline\hline
     \multirow{6}{*}{Hybrid} & \multirow{2}{*}{RWTH\cite{luscher2019rwth}} & 4g & 3.8 & 8.8  \\
        &  & +NNLM & 2.3 & 5.0 \\
        \cline{2-5}
        & \multirow{2}{*}{Han et al.\cite{han2019state}} & 4g & 2.9 & 8.3 \\
        &  & +NNLM & \textbf{2.2} & 5.8 \\
        \cline{2-5}
        & Le et al.\cite{le2019senones} & 4g & 3.2 & 7.6 \\
        \cline{2-5}
        & \multirow{2}{*}{Ours} & 4g & \textbf{2.60} & \textbf{5.59} \\
        &                       & +NNLM & 2.26 & \textbf{4.85} \\
    \hline
    \end{tabular}
    
    \label{tab:comp}
\end{table}
\vspace{-0.5em}

\subsection{Limited Right Context}
\vspace{-0.5em}
All the transformer-based experiments so far used full context. To understand to what extent the transformer relies on future frames to derive embeddings for the current frames, we take the \textit{vggTrf(768, 12)} model (row 4, Table \ref{tab:arch}) and force every layer to attend to a fixed limited right context during inference. Interestingly,  though this creates a large mismatch between training and inference, the resultant systems can still yield reasonable WERs if the number of right context frames is large enough. Note that though each layer only requires limited right context frames, the overall right context length is added up by the right context length of every transformer layer, therefore we still end up with a large look-ahead window into the future, which makes it less possible to be used in a streaming ASR application. We will investigate transformer-based acoustic models with the streaming constraint in our future study.

\begin{table}[tb]
    \centering
    \caption{Forcing transformer models to use limited right context (RC) \emph{per layer} during inference. Given a 12-layer transformer, an RC of 10 frames translates to 2.48 seconds of total lookahead.}
    \begin{tabular}{|c||cc|}
    \hline
    RC & test-clean & test-other \\ 
    \hline
    $\infty$ & 2.87 & 6.45 \\
    50       & 3.01 & 7.12 \\
    20       & 3.29 & 8.10 \\
    10       & 3.65 & 9.01 \\
    \hline
    \end{tabular}
    \label{tab:rc}
\end{table}
\vspace{-0.5em}

\subsection{Large Scale Experiments}
\vspace{-0.5em}
Finally, we perform a large scale experiment on one of our internal tasks, \emph{English video ASR}. The training set consists of 13.7K hours of videos (from 941.6K video clips) shared publicly by users; only the audio part of those videos are used in our experiments. These data are completely anonymized; both transcribers and researchers do not have access to any user-identifiable information. Due to the data nature, it is a very diverse and challenging task.  About 9 hours (from 620 video clips) data are held out for \texttt{dev} set. 3 test sets are used for evaluation purpose: an 8.5-hour \texttt{curated} set of carefully select very clean videos, an 19-hour \texttt{clean} set and a 18.6-hour \texttt{noisy} set. For our initial evaluation purpose, both training and test sets are segmented into maximum 10 second segments. 

Due to time limit, we only built \textit{vggTrf(768, 12)} without the iterated loss and \textit{vggBLSTM(800, 5)} on this task. Table \ref{tab:video_asr} shows that on this task, the proposed transformer-based acoustic model outperform vggBLSTM by 4.0-7.6\%. We will report more results in our future work.  
\vspace{-1em}

\begin{table}[htb]
    \centering
    \caption{Experiment results on our internal \emph{English video ASR} task.}
    \begin{tabular}{|c|ccc|}
    \hline
    Model     & curated & clean & noisy  \\
    \hline\hline
    vggBLSTM(800,5) & 10.72 & 15.97 & 22.13 \\
    vggTrf(768,12) &  9.90 &  15.26 & 21.25 \\
    \hline
    \end{tabular}
    \label{tab:video_asr}
\end{table}
\vspace{-1em}


\section{Discussions And Conclusions}
\label{sec:con}

In this work, we proposed and evaluated transformer-based acoustic models for hybrid speech recognition. A couple of model modeling choices are discussed and compared. We demonstrated that transformer can significantly outperforms BLSTM and give the best acoustic models on Librispeech benchmark. Initial study on a much larger and more challenging dataset also confirms our findings. 

There are many works we are yet to explore. For example, our experiments did not show to what extent transformer's superior performance comes from replacing recurrence with self-attention, while other modeling techniques from transformer can be borrowed to improve RNNs as well \cite{chen2018best}. The quadratically growing cost with respect to the length of speech signals is still a major blocker for transformer-based acoustic models to be used in practice. These questions will be studied in our future work. 

\footnotesize
\bibliographystyle{IEEEbib}
\bibliography{strings,refs}

\end{document}